\pdfoutput=1

\documentclass[11pt]{article}

\usepackage{emnlp2021}

\usepackage{times}
\usepackage{latexsym}

\usepackage[T1]{fontenc}

\usepackage[utf8]{inputenc}

\usepackage{microtype}

\usepackage{amsmath}
\usepackage{graphicx}
\usepackage{float}
\usepackage{subcaption}
\usepackage{multicol}
\usepackage{booktabs}
\usepackage{multirow}
\usepackage[bottom]{footmisc}
\usepackage{cancel}
\usepackage{caption}
\usepackage{subcaption}
\usepackage{pifont}
\newcommand{\xmark}{\ding{55}}%
\usepackage{enumerate}

\title{Sparsity and Sentence Structure\\in Encoder-Decoder Attention of Summarization Systems}

\author{Potsawee Manakul \textnormal{ and } Mark J. F. Gales \\
  Department of Engineering, University of Cambridge \\
  \texttt{pm574@cam.ac.uk, mjfg@eng.cam.ac.uk}}

\begin{document}
\maketitle
\begin{abstract}
Transformer models have achieved state-of-the-art results in a wide range of NLP tasks including summarization. Training and inference using large transformer models can be computationally expensive. Previous work has focused on one important bottleneck, the quadratic self-attention mechanism in the encoder. Modified encoder architectures such as LED or LoBART use local attention patterns to address this problem for summarization. In contrast, this work focuses on the transformer's encoder-decoder attention mechanism. The cost of this attention becomes more significant in inference or training approaches that require model-generated histories. First, we examine the complexity of the encoder-decoder attention. We demonstrate empirically that there is a sparse sentence structure in document summarization that can be exploited by constraining the attention mechanism to a subset of input sentences, whilst maintaining system performance. Second, we propose a modified architecture that selects the subset of sentences to constrain the encoder-decoder attention. Experiments are carried out on abstractive summarization tasks, including CNN/DailyMail, XSum, Spotify Podcast, and arXiv.\footnote{Our code is available at \url{https://github.com/potsawee/encdec_attn_sparse}.}
\end{abstract}

\section{Introduction}
The Transformer architecture \cite{vaswani2017attention} with large-scale pre-training has become the de-facto approach for a wide range of NLP tasks, from classification \cite{devlin-etal-2019-bert} to seq2seq \cite{raffel2020exploring}. Training and inference using large transformer models can be computationally expensive because the self-attention's time and memory grow quadratically with sequence length. Hence, there has been significant interest in efficient transformer architectures. A number of approaches have been proposed to tackle the quadratic complexity, and a comprehensive survey on efficient transformers has been compiled in \citet{tay2020efficient}. Most existing approaches are developed for encoder-only architectures. For seq2seq tasks, efficient models such as BigBird \cite{zaheer2020big} or LED \cite{beltagy2020longformer} consist of an efficient encoder with the vanilla decoder. For long-document summarization, this combination has been shown effective because the major bottleneck is the encoder self-attention \cite{manakul2021_longspan}. The attention mechanisms in the decoder consist of self-attention and encoder-decoder attention. Techniques such as local attention are applicable to self-attention in both the encoder and decoder, while this work focuses on the encoder-decoder attention.

When humans produce a summary, the information conveyed by each word/part in the summary is likely drawn from some key sentences in the original document. Inspired by this, we hypothesize that if the encoder-decoder attention is constrained dynamically to salient sentences, the computation cost will be reduced. For instance, sentence-level structures for the encoder-decoder attention have been shown effective in the traditional RNN encoder-decoder attention \cite{cohan-etal-2018-discourse, li-etal-2019-keep, manakul2020_interspeech}

In this work, first, we compare the decoder's cost in the training and inference stages. We study the sparsity of the encoder-decoder attention in a common transformer-based abstractive summarization model. An approximation method to exploit this sparsity is described, and an empirical upper bound performance is given. Second, we propose a modified decoder architecture that can dynamically select salient input sentences to constrain the encoder-decoder attention without having to compute complete attention at inference time. Techniques to train our proposed model are described, and compared to the full attention baseline performance  and empirical upper bound.

\section{Models and Data}
\textbf{Vanilla Transformers.} We use BART \cite{lewis-etal-2020-bart} and local-attention BART (LoBART) \cite{manakul2021_longspan} as our base models. BART's maximum input length is 1024, while that of LoBART is 4096 with attention width of 1024. BART is fine-tuned to CNN/DailyMail and XSum, and LoBART is fine-tuned to Podcast and arXiv.

\vspace{4pt}
\noindent \textbf{Data.} CNN/DailyMail \cite{hermann2015teaching} and XSum \cite{narayan-etal-2018-dont} are used with BART, while long-document arXiv \cite{cohan-etal-2018-discourse} and Spotify Podcast \cite{clifton-etal-2020-100000} are used with LoBART. More details about models, data, and training are provided in Appendix \ref{sec:reproducibility}.

\section{Attention in the Transformer}
\label{sec:encdec_attn}
Time and memory are dominated by the encoder self-attention, and models such as LoBART adopt local attention in its encoder to mitigate this bottleneck, while keeping the original decoder \cite{manakul2021_longspan}. Training is fast because attention is highly parallelizable. However, during inference, the decoder uses its histories, becoming less parallelizable. To understand when the decoder might become a bottleneck, we fix the input length $N$ and measure the computational time as a function of the target length $M$: 
\begin{equation}
    {\tt time}=\bar{c}_1+\bar{c}_2M+\bar{c}_3M^2 \label{eq:model_cost_with_M}
\end{equation}
in three operating modes: i) Forward+Backward, e.g. at training time; ii) Forward only, e.g. forward-pass where the input to the decoder is provided in advance; iii) Inference, e.g. the decoder using its own back histories as the input.

Through a curve-fitting method, the results in Table \ref{tab:decoder_cost} show that the relative decoder cost during inference is almost one order of magnitude larger than that during training, e.g. forward+backward or forward only. More details are provided in Appendix \ref{sec:time_appendix}, where we also show that the encoder-decoder attention cost is greater than the decoder self-attention cost. Therefore, this work will focus on the encoder-decoder attention.


\begin{table}[!h]
  \centering
 \tabcolsep=0.09cm
  \begin{tabular}{rcc}
    \toprule
    Mode  &$\bar{c}_2/\bar{c}_1$ ($10^{-3}$) &$\bar{c}_3/\bar{c}_1$ ($10^{-6}$) \\
    \midrule
    Forward+Backward  &1.08 &0.17 \\
    Forward only          &1.14 &0.25 \\
    Inference         &9.96 &1.30 \\
    \bottomrule
  \end{tabular}
  \caption{Empirical computational time as a function of the target length $M$ where $\bar{c}_1, \bar{c}_2, \bar{c}_3$ are the coefficients in Eq. \ref{eq:model_cost_with_M}. The analysis is based on BART from \citet{wolf-etal-2020-transformers} and the input length is 1024.
  }
  \label{tab:decoder_cost}
\end{table}

\subsection{Encoder-Decoder Attention}
\label{sec:sparsity_intro}
Let $M$ = the summary length, $N$ = the input length, $N_1$ = \#sentences, and $N_2$ = the average number of words in a sentence, e.g. $N = N_1N_2$. The standard encoder-decoder attention in Eq. \ref{eq:standard_attention} (scaling factor omitted) where $\mathbf{Q} \in \mathcal{R}^{M \times D}$ and $\mathbf{K},\mathbf{V} \in \mathcal{R}^{N\times D}$ has the complexity: $\mathcal{O}(MN)=\mathcal{O}(MN_1N_2)$. Note that we fix the representation dimension $D$, so $D$ is omitted in our complexity notation.
\begin{equation}
 \mathbf{A} = \text{softmax}(\mathbf{Q}\mathbf{K}^T)\mathbf{V} \label{eq:standard_attention}
\end{equation}
If the attention is concentrated on some $r$ sentences,\footnote{Motivated by the observations shown in Appendix \ref{sec:attention_plot_appendix}.} by selecting appropriate $r$, the speed of the encoder-decoder attention can be improved by a factor of $N_1/r$ in average. This is equivalent to:
\begin{equation}
 {\mathbf{A}} \approx \hat{\mathbf{A}} = \text{softmax}(\mathbf{Q}\hat{\mathbf{K}}^T)\hat{\mathbf{V}}
 \label{eq:subset_attention_hat}
\end{equation}
where $\hat{\mathbf{K}},\hat{\mathbf{V}} \in \mathcal{R}^{rN_2 \times D}$, resulting in $\mathcal{O}(MrN_2)$. 
\subsection{Sparsity of Encoder-Decoder Attention}
\label{section:sparisity_encdec_attn}
\begin{figure}[H]
    \begin{subfigure}[b]{0.49\linewidth}
    \centering
      \includegraphics[width=0.99\linewidth,keepaspectratio]{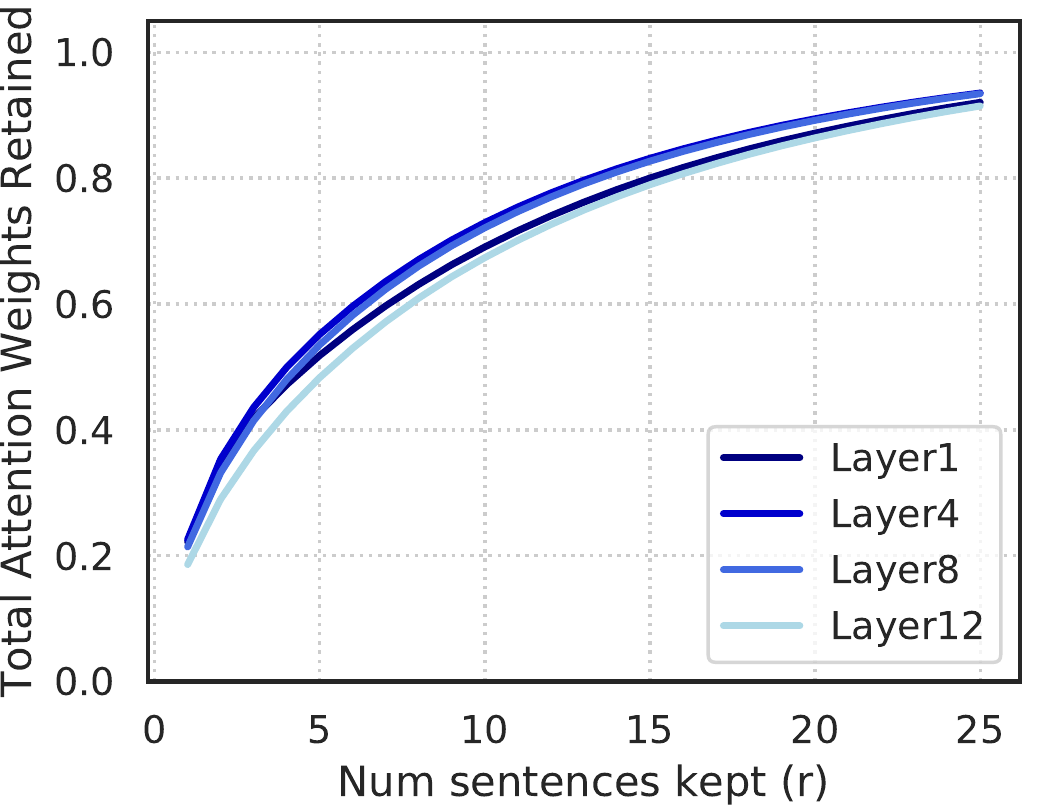}
    \caption{Layers 1,4,8,12}
    \label{fig:attention_retained_vanilla} 
    \end{subfigure}
        \hfill
    \begin{subfigure}[b]{0.49\linewidth}
    \centering
      \includegraphics[width=0.99\linewidth,keepaspectratio]{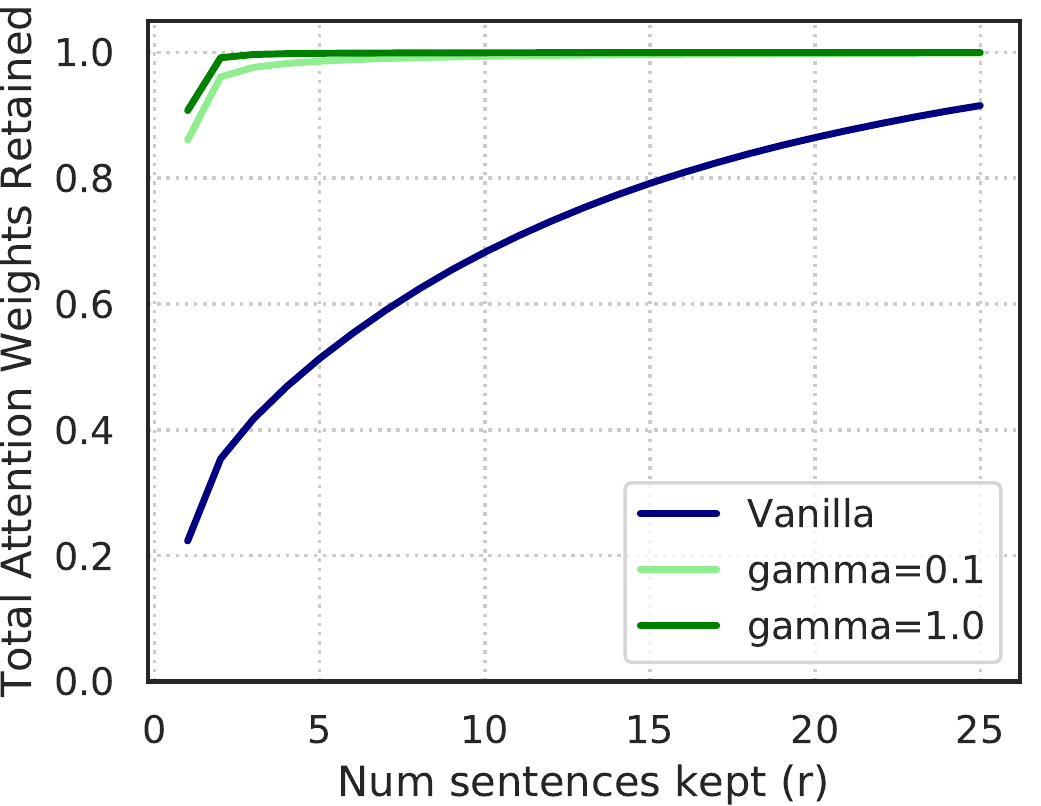}
    \caption{$\mathcal{L}_{\tt A}$-tuned (Layer1)}
    \label{fig:attention_retained_tuned} 
    \end{subfigure}
    \caption{The sum of attention weights against the number of retained sentences ($r$) evaluated on CNNDM.}
    \label{fig:attention_retained}    
\end{figure}

Let the subscript $(i,j)$ denote the position of the $j$-th word in the $i$-th input sentence, e.g. $\mathbf{K} = [\underbrace{\mathbf{k}_{1,1},\mathbf{k}_{1,2},\mathbf{k}_{1,J_1}}_{\text{sent} 1},...,\underbrace{\mathbf{k}_{i,1},\mathbf{k}_{i,J_i}}_{\text{sent} i},...,\underbrace{\mathbf{k}_{N_1,1},\mathbf{k}_{N_1,J_{N_1}}}_{\text{sent} N_1}]$. At inference time, the outputs are generated sequentially: $\mathbf{a}_m$$=$$\text{softmax}(\mathbf{q}_m\mathbf{K}^T)\mathbf{V}$, so $r$ sentences can be determined independently for each $\mathbf{q}_m$. Consider the following sum of attention weights as the saliency at decoding step $m$ of sentence $i$:\footnote{We discuss the details of \textit{multi-head} attention on $\alpha^{\tt s}_{m,i}$ and other operations such as entropy in Appendix \ref{section:multihead_appendix}}
\begin{equation}
  \alpha^{\tt s}_{m,i} = \frac{1}{Z_m} \sum_{j=1}^{J_i} \exp (\mathbf{q}_m \cdot \mathbf{k}_{i,j})
    \label{eq:sent_level_attn}
\end{equation}
where $Z_m=\sum_{\forall i'} \sum_{\forall j'} \exp (\mathbf{q}_m \cdot \mathbf{k}_{i',j'})$. We then compute $\sum_i \alpha^{\tt s}_{m,i}$ up to $r$ sentences ranked by $\alpha^{\tt s}_{m,i}$. The results in Fig. \ref{fig:attention_retained_vanilla} show that $r$=25 is required to achieve the sum of attention weights at 90\%. In addition to the vanilla model, we can fine-tune BART explicitly to make the attention sparse using:
\begin{equation}
\mathcal{L}_{\tt A} = \mathcal{L}_{{\tt xent}} + \gamma\mathcal{L}_{{\tt sparse}}
\end{equation}
where $\mathcal{L}_{{\tt xent}}$ is the teacher-forced cross entropy loss, $\mathcal{L}_{{\tt sparse}} = \frac{1}{M}\sum_{m=1}^M \text{H} ( \boldsymbol{\alpha}^{{\tt s}}_m )$, and entropy $\text{H}(\boldsymbol{\alpha}^{{\tt s}}_m) = -\sum_{i=1}^{N_1}{\alpha}^{{\tt s}}_{m,i} \log {\alpha}^{{\tt s}}_{m,i}$. We show in Fig. \ref{fig:attention_retained_tuned} that the fine-tuned models ($\gamma$=0.1 \& $\gamma$=1.0) retain close to 100\% of attention weights for small $r$. Subsequently, we investigate how selecting $r$ sentences impacts the summarization performance.

To obtain an empirical upper bound performance of Eq. \ref{eq:subset_attention_hat}, for each $\mathbf{q}_m$, we can get \textit{ideal} $\mathbf{k},\mathbf{v}$ corresponding to the top $r$ sentences ranked by $\alpha^{\tt s}_{m,i}$:
\begin{equation}
\mathcal{I}^r_m=[(i,j) \hspace{0.125cm}\text{s.t.}\hspace{0.125cm} i \in \text{top-}r(\alpha^{\tt s}_{m,i})] \label{eq:idealI}
\end{equation}
$\hat{\mathbf{K}}_m = [\mathbf{k}_{i,j}: (i,j) \in \mathcal{I}^r_m]$, and the same method is applied to obtain $\hat{\mathbf{V}}_m$.

\begin{table}[!h]
  \centering
  \tabcolsep=0.14cm
  \begin{tabular}{c|cc|ccc}
    \toprule
    System          &$r$ &$\mathcal{I}^r_m$   &R1 &R2 &RL  \\
    \midrule
     Vanilla        &All &N/A    &44.03 &20.92 &40.99  \\
     ($\gamma=0.0$) &5 &Ideal    &43.94 &20.82 &40.81  \\
                    &5 &Random    &39.06 &14.32 &36.07  \\
    \midrule
    $\gamma=0.1$    &5 &Ideal    &44.22 &21.01 &41.19 \\
    $\gamma=1.0$    &5 &Ideal    &43.61 &20.46 &40.60 \\

    \bottomrule
  \end{tabular}
  \caption{Sparsity and Selection ($\mathcal{I}^r_m$) on CNNDM.}
  \label{tab:expA_results}
\end{table}
\noindent The results in Table \ref{tab:expA_results} show that: 
\begin{itemize}
    \item For the vanilla model, despite the sum of attention weights being around 50\% at $r$=5 (Fig. \ref{fig:attention_retained_vanilla}), the model is sufficiently sparse, and constraining to $r$ ideal sentences ($\text{All} \rightarrow$ $\mathcal{I}^{r,\text{Ideal}}_m$) results in a small performance degradation.
    \item Forcing for sparsity (Fig.\ref{fig:attention_retained_tuned}) does \textit{not} yield a significant performance improvement; but this forcing also makes the model more sensitive to random selection (results in Appendix \ref{section:random_selection_appendix}).
\end{itemize}
Thus, for summarization, there is an observable sparsity, which allows us to reduce the cost of encoder-decoder attention with a minimal degradation. Next, we investigate how to build an efficient form of approximator to obtain salient sentences.

\begin{figure*}[!t]
    \begin{subfigure}[b]{0.245\linewidth}
    \centering
      \includegraphics[width=\linewidth,keepaspectratio]{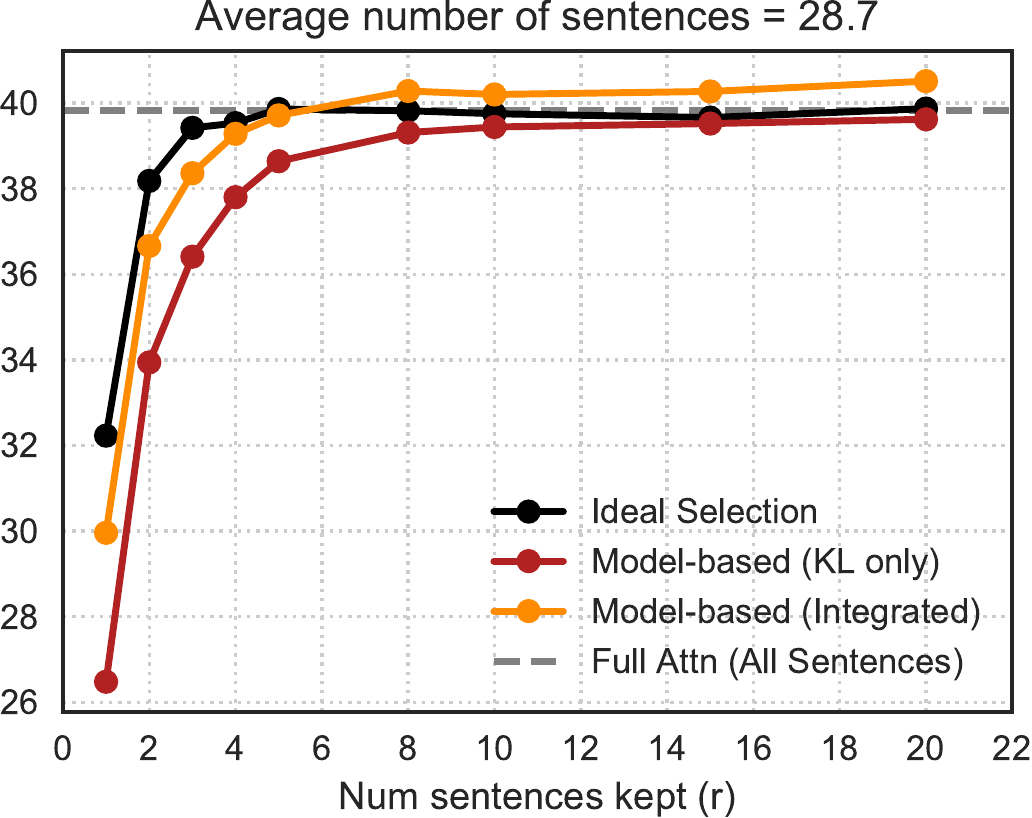}
    \caption{BART(1k) \& CNNDM}
    \label{fig:main_cnndm}  
    \end{subfigure}
        \hfill
    \begin{subfigure}[b]{0.245\linewidth}
    \centering
      \includegraphics[width=\linewidth,keepaspectratio]{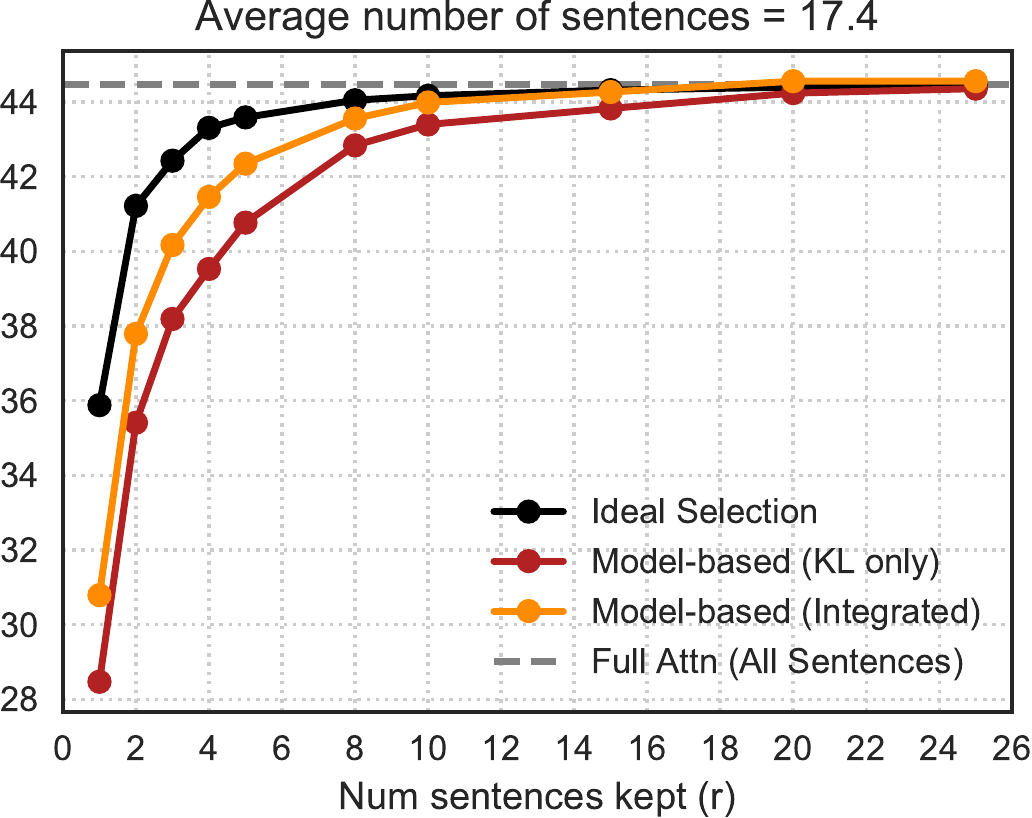}
    \caption{BART(1k) \& XSum}
    \label{fig:main_xsum} 
    \end{subfigure}
        \hfill
    \begin{subfigure}[b]{0.245\linewidth}
    \centering
      \includegraphics[width=\linewidth,keepaspectratio]{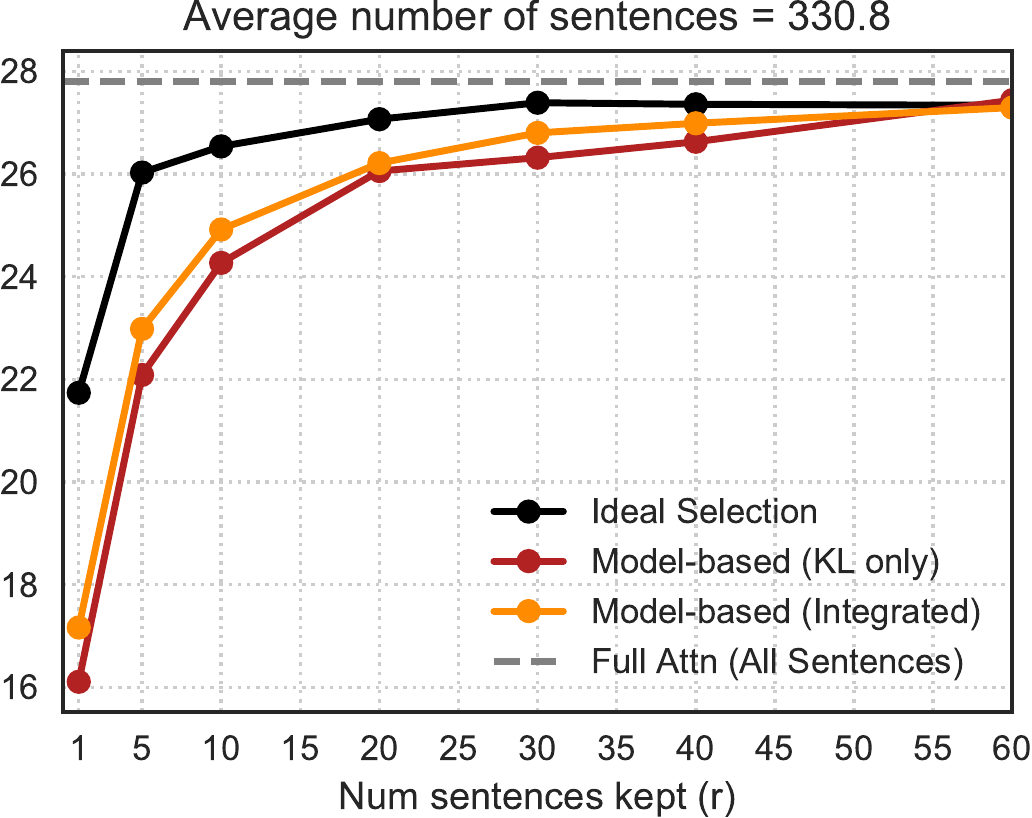}
    \caption{LoBART(4k) \& Podcast}
    \label{fig:main_podcast} 
    \end{subfigure}
        \hfill
    \begin{subfigure}[b]{0.245\linewidth}
    \centering
      \includegraphics[width=\linewidth,keepaspectratio]{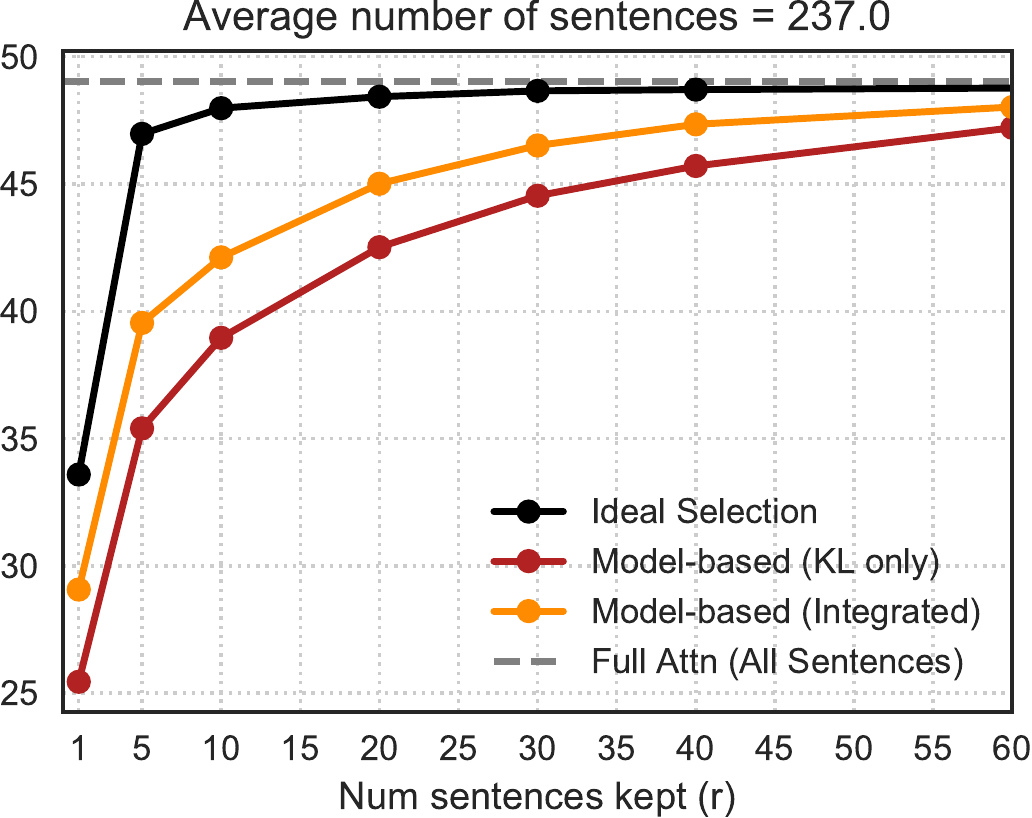}
    \caption{LoBART(4k) \& arXiv}
    \label{fig:main_arxiv} 
    \end{subfigure}
    \caption{Performance (ROUGE-1) of BART \& LoBART. The integrated training is based on $\mathcal{I}^{r,\text{Apx}}_m$.}
    \label{fig:summary_performance}    
\end{figure*}

\section{Sentence-Level Structure for Encoder-Decoder Attention}
\label{sec:sent_level_structure}
In Section \ref{sec:sparsity_intro}, we use ideal selection $\mathcal{I}^r_m$ (Eq. \ref{eq:idealI}), which requires computing $\alpha^{\tt s}_{m,i}$ (Eq. \ref{eq:sent_level_attn}) using all input words. This process cannot make the decoder more efficient. By exploiting the sentence structure in the document, we propose the following partition for the sentence-level attention score (Eq. \ref{eq:sent_level_attn}) to allow a compact approximation:
\begin{equation}
\resizebox{.99\hsize}{!}{$
{\alpha}^{\tt s}_{m,i} \approx \tilde{\alpha}^{\tt s}_{m,i} = \text{softmax}\left(f_1({\mathbf{q}_m}) \cdot f_2 ({\mathbf{k}_{i,1}, ..., \mathbf{k}_{i,J_i}})\right)
$} 
\label{eq:sent_level_approximation}
\end{equation}
where $\sum_{i=1}^{N_1}\tilde{\alpha}^{\tt s}_{m,i} = 1.0$. Essentially, we modify the standard encoder-decoder attention such that it performs sentence selection based on $\tilde{\alpha}^{\tt s}_{m,i}$ (Eq. \ref{eq:sent_level_approximation}) and computes subset attention $\hat{\mathbf{A}}$ (Eq. \ref{eq:subset_attention_hat}).

\subsection{Complexity of Modified Attention}
The modified encoder-decoder attention consists of two components: i) sentence-level attention over $N_1$ sentences; ii) word-level attention over $rN_2$ words. Let $p$ denote a unit of matrix multiplication cost and $q$ denote a unit of softmax cost. The costs associated with attention are:
\begin{enumerate}[i)]
    \item Sentence-level (Eq.\ref{eq:sent_level_approximation}): $pMN_1D + qMN_1$
    \item Word-level (Eq.\ref{eq:subset_attention_hat}): $2pMrN_2D + qMrN_2$
\end{enumerate}
The additional cost associated with the sentence-level representation on the encoder side grows with the input length $N$=$N_1N_2$. Thus, as opposed to $\mathcal{O}(MN_1N_2)$ in the case of vanilla encoder-decoder attention, the overall complexity of the modified attention is $\mathcal{O}(MN_1 + k_wMrN_2 + k_eN_1N_2)$, where $k_w \approx \frac{2pD+q}{pD+q}$ and $k_e$ depends on the exact form of sentence-level representation computation.


\subsection{Model-based Neural Approximator}
\label{section:model_based_neural_model}
\begin{figure}[H]
    \centering
      \includegraphics[width=0.99\linewidth,keepaspectratio]{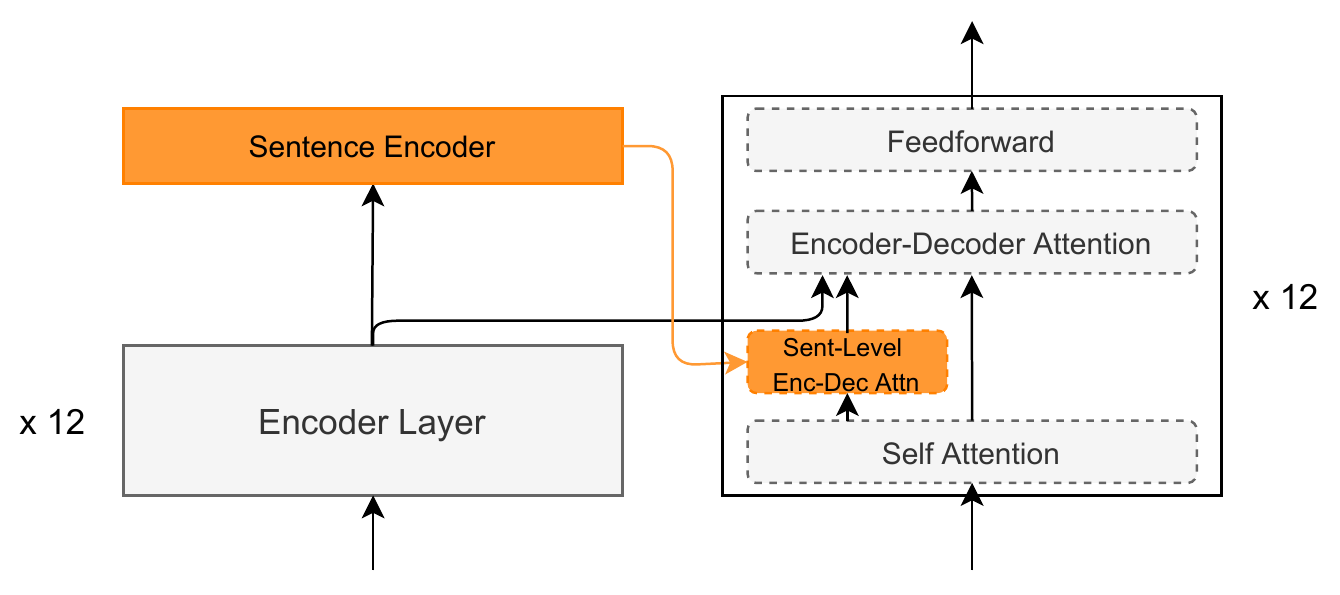}
    \caption{Modified architecture with model-based approximator where the base model is BART/LoBART. Model-based neural approximator is shown in orange.}
    \label{fig:modified_architecture}    
\end{figure}

To utilize the simple partition and sentence-level structure in Eq. \ref{eq:sent_level_approximation}, we use a linear mapping for $f_1$ and a bidirectional RNN for $f_2$ as follows:
\begin{align}
    f_1(\mathbf{q}_m) &= \mathbf{q}_m \mathbf{W}^{\tt Q}  \label{eq:query_mapping} \\
f_2(\mathbf{k}_{i,1},...,\mathbf{k}_{i,J_i}) &= \mathbf{y}_i \mathbf{W}^{\tt K} \label{eq:key_mapping} \\
\mathbf{y}_i = \text{RNN}(\mathbf{k}&_{i,1},...,\mathbf{k}_{i,J_i})\label{eq:sentence_level_representation}
\end{align}
As illustrated in Fig. \ref{fig:modified_architecture}, the base transformer model is extended by augmenting two layers: i) sentence-level encoder-decoder attention computing $\tilde{\alpha}^{\tt s}_{m,i}$ in Eq. \ref{eq:sent_level_approximation}; ii) sentence encoder computing the sentence-level representation in Eq. \ref{eq:sentence_level_representation}. The details about model parameters are provided in Appendix \ref{sec:reproducibility}.
\vspace{-0.3cm}
\subsection{KL Loss and Integrated Training}
Let $\boldsymbol{\theta}_\text{dec}$ denote the original decoder, and $\tilde{\boldsymbol{\theta}}$ denote the neural approximator. We train $\tilde{\boldsymbol{\theta}}$ by minimizing:
\vspace{-0.2cm}
\begin{equation}
    \mathcal{L}_{\tt KL} = \frac{1}{M}\sum_{m=1}^M  \text{KL}\left( \boldsymbol{\alpha}_{m}^{{\tt s}}||\boldsymbol{\tilde{\alpha}}_{m}^{{\tt s}} \right) 
\vspace{-0.05cm}
\end{equation}
where $\text{KL}(.)=\sum_{i=1}^{N_1}{\alpha}_{m,i}^{{\tt s}} \log ({{\alpha}_{m,i}^{{\tt s}}}/{{\tilde{\alpha}}_{m,i}^{{\tt s}}})$. In addition, we can integrate $\boldsymbol{\theta}_\text{dec}$ in the training process. With the teacher-forced cross entropy loss $\mathcal{L}_{\tt xent}$, we define the \textbf{integrated training} loss:
\begin{equation}
    \mathcal{L}_{\tt I} =  \mathcal{L}_{\tt{xent}} + \lambda  \mathcal{L}_{\tt{KL}}
\end{equation}
We cannot optimize $\mathcal{L}_{\tt I}$ in an end-to-end fashion because the top-$r$ operation in Eq. \ref{eq:idealI} is not differentiable. Hence, we interleave the training, i.e. update $\boldsymbol{\theta}_\text{dec}$ at fixed $\tilde{\boldsymbol{\theta}}$ and update $\tilde{\boldsymbol{\theta}}$ using $\mathcal{L}_{\tt KL}$ only:
\vspace{-0.5cm}
\begin{equation}
\resizebox{.99\hsize}{!}{$
     \Delta\boldsymbol{\theta}_{\text{dec}} =\nabla_{\boldsymbol{\theta}_{\text{dec}}}\mathcal{L}_{\tt I}|_{\tilde{\boldsymbol{\theta}}} =  \nabla_{\boldsymbol{\theta}_{\text{dec}}}\mathcal{L}_{\tt{xent}}|_{\tilde{\boldsymbol{\theta}}} + \lambda \nabla_{\boldsymbol{\theta}_{\text{dec}}} \mathcal{L}_{\tt{KL}}|_{\tilde{\boldsymbol{\theta}}}
     \label{eq:integrated_dec}
$}
\end{equation}
\vspace{-1cm}
\begin{equation}
      \Delta\tilde{\boldsymbol{\theta}} = \nabla_{\tilde{\boldsymbol{\theta}}}\mathcal{L}_{\tt I} =  \cancelto{0}{\nabla_{\tilde{\boldsymbol{\theta}}}\mathcal{L}_{\tt{xent}}} + \lambda \nabla_{\tilde{\boldsymbol{\theta}}} \mathcal{L}_{\tt{KL}} 
      \label{eq:integrated_selector} 
\end{equation}
Because during training, we compute both ${\alpha}^{\tt s}_{m,i}$ (ideal) and $\tilde{\alpha}^{\tt s}_{m,i}$ (approx), we can use either in the top-$r$ selection. Also, inspired by scheduled sampling \cite{scheduled_bengio}, we try mixing them: ${\alpha}^{\tt s}_{m,i}$ with probability $1-\frac{\text{step}}{\text{epoch\_size}}$, otherwise $\tilde{\alpha}^{\tt s}_{m,i}$.
\subsection{System Performance}
\label{section:system_performance}
\begin{table}[!h]
  \centering
  \tabcolsep=0.05cm
  \begin{tabular}{c|cc|ccc}
    \toprule
    System  &Train &Inference     &R1 &R2 &RL   \\
    \midrule
    Vanilla &\xmark   &$\mathcal{I}_m^{r\text{,Rnd}}$  &39.06 &14.32 &36.07 \\
    Vanilla &\xmark   &$\mathcal{I}_m^{r\text{,Idl}}$  &43.94 &20.82 &40.81  \\
    \midrule
    KL-only  &$\mathcal{L}_{\tt KL}$       &$\mathcal{I}_m^{r\text{,Apx}}$
    &43.02 &20.02 &39.89 \\
    Int-Idl &$\mathcal{L}_{\tt I}$($\mathcal{I}_m^{r\text{,Idl}}$)  &$\mathcal{I}_m^{r\text{,Apx}}$ &43.03 &20.04 &40.05 \\ 
    Int-Apx &$\mathcal{L}_{\tt I}$($\mathcal{I}_m^{r\text{,Apx}}$)  &$\mathcal{I}_m^{r\text{,Apx}}$ &43.72 &20.40 &40.70 \\ 
    Int-Mix &$\mathcal{L}_{\tt I}$(Mix)    &$\mathcal{I}_m^{r\text{,Apx}}$ &43.31 &20.21 &40.35 \\ 
    \bottomrule
  \end{tabular}
  \caption{Performance on CNNDM where $r$=5 for both training and inference. KL-only = $\tilde{\boldsymbol{\theta}}$ trained on $\mathcal{L}_{\tt KL}$; Int = ($\boldsymbol{\theta}_\text{dec}$\&$\tilde{\boldsymbol{\theta}}$) trained on $\mathcal{L}_{\tt I}$. Rnd=random, Idl=ideal, Apx=approximation, Mix=scheduled(Idl/Apx).}
  \label{tab:integrated_training}
\end{table}

In Table \ref{tab:integrated_training}, we provide two vanilla baselines: \textit{random} ($\mathcal{I}_m^{r\text{,Rnd}}$) obtained by random selection; \textit{ideal} ($\mathcal{I}_m^{r\text{,Idl}}$) obtained by Eq. \ref{eq:idealI}. The results show that the KL-only system clearly outperforms the random selection baseline, and the performance degradation of the KL-only system can be reduced by our integrated training. The results verify the \textit{effectiveness} of our modified decoder that attends to a subset of sentences. Also, Table \ref{tab:integrated_training} shows that it is best to use $\mathcal{I}_m^{r\text{,Apx}}$ as reference in integrated training. This result is likely because we initialized integrated training from the KL-only model. In addition, we apply the modified architecture to BART trained on XSum ($\leq$1k words), and to LoBART trained on Podcast and arXiv ($\leq$4k words). The results in Fig. \ref{fig:summary_performance} confirm that the performance of our proposed method converges to that of the full attention baseline across all models and datasets.

In addition, $r^*\hspace{-5pt} \approx \hspace{4pt} $5,10,30,30, respectively.\footnote{$r^*$ denotes $r$ at which the ideal selection system's performance plateaus/reaches the full attn baseline's performance.} Although XSum has fewer sentences in average compared to CNNDM, $r^*_{\text{XSum}}$$>$$r^*_{\text{CNNDM}}$ as XSum is more abstractive. For longer summarization tasks as shown by Podcast and arXiv, the performance degradation appears larger, meaning that the task of constraining to salient sentences in longer tasks is more challenging, and larger $r$ is required.

\subsubsection*{Sensitivity of $r$ in Integrated Training}
\begin{figure}[!h]
    \begin{subfigure}[b]{0.49\linewidth}
    \centering
      \includegraphics[width=0.99\linewidth,keepaspectratio]{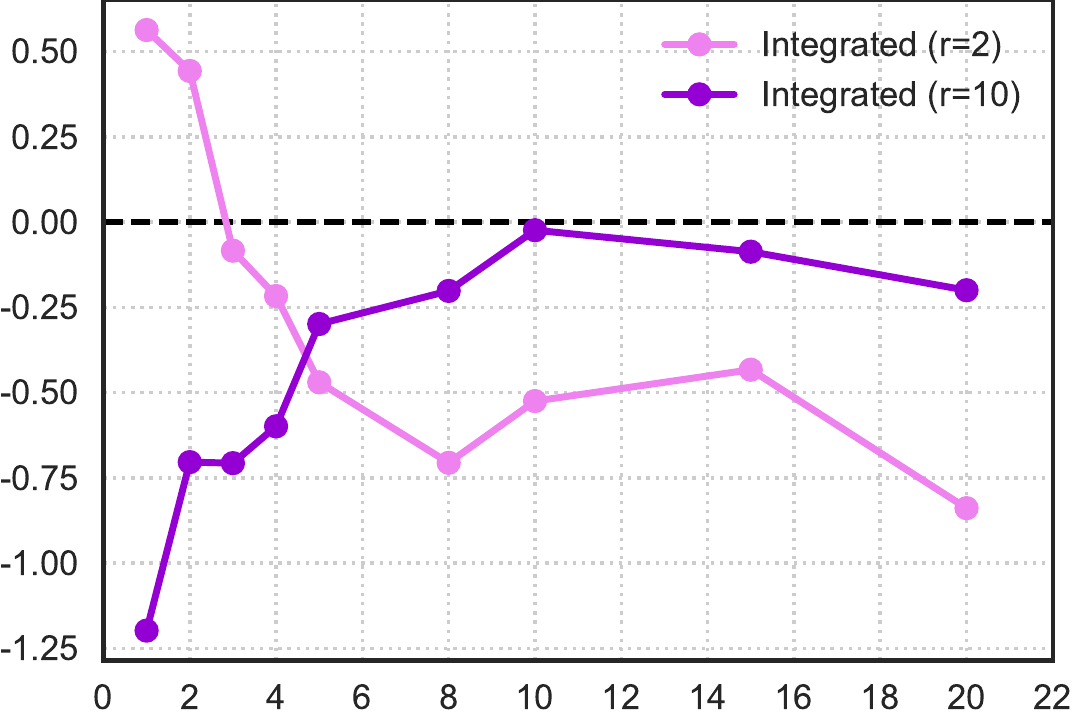}
    \caption{CNNDM}
    \end{subfigure}
        \hfill
    \begin{subfigure}[b]{0.49\linewidth}
    \centering
      \includegraphics[width=0.99\linewidth,keepaspectratio]{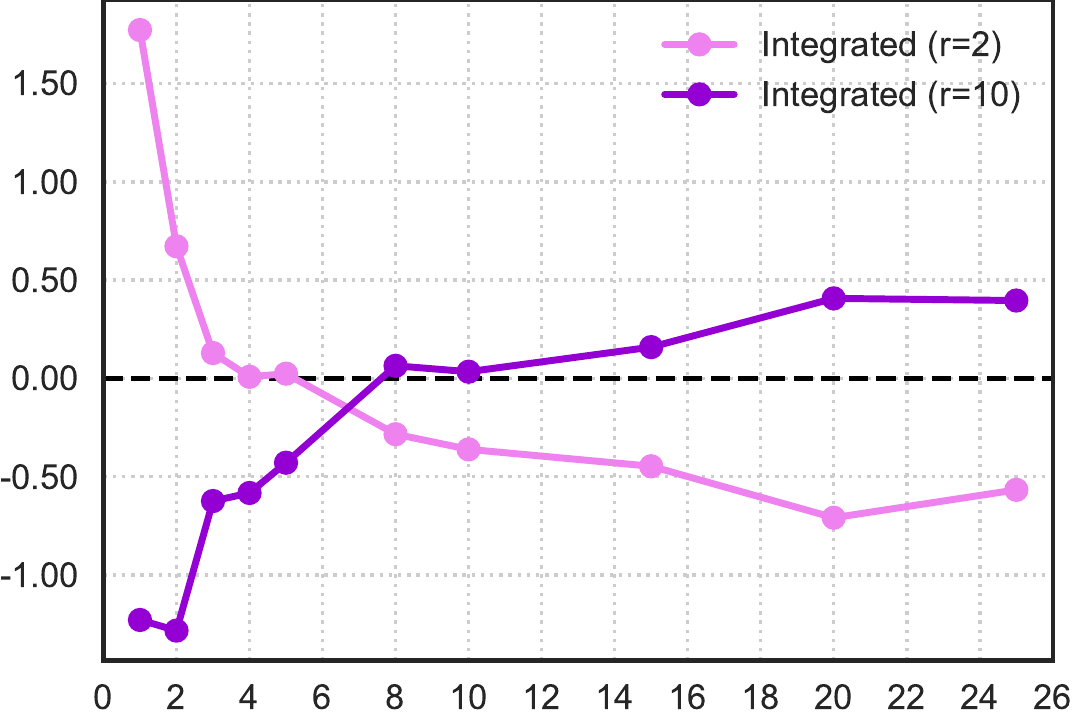}
    \caption{XSum}
    \end{subfigure}
    \caption{$\Delta$R1 (Y-axis) against $r$ at inference (X-axis).}
    \label{fig:varying_r}    
\end{figure}

We train BART in three settings: $r^{\text{train}}$=2,5,10, and we show the performance level w.r.t. the model with $r^{\text{train}}$=5 in Fig. \ref{fig:varying_r}. The results show that setting $r^{\text{train}}$ beyond $r^*$ is not necessarily beneficial as shown by the model with $r^{\text{train}}$=10 in the CNNDM result, and it is best to set $r^{\text{train}}$ close to $r^{\text{inference}}$.

\subsection{Further Discussion on Sentence-Level Encoder-Decoder Attention}
The results in Section \ref{section:system_performance} demonstrate empirically that a neural network can predict sparsity, therefore, allowing sentence selection. Our novel framework requires an addition of modules to the original attention mechanism, and the real gain in speed will depend on the balance of the sparsity against the computational cost of the additional modules. Consequently, the challenge is to make these additional modules highly efficient. Because the particular network realization selected in this paper is to show the feasibility of our framework, its limitations and possible improvements are discussed as follows.     
\subsubsection*{Limitations and Possible Improvements}
The model choice of using RNN for the sentence encoder in Eq. \ref{eq:sentence_level_representation} leads to a large additional computational cost (specifically large $k_e$) because the computational cost of RNN grows with $N_1N_2D^2$. 
Because the goal is to obtain a sentence-level representation, there is an opportunity to replace RNN by a hierarchical attention that runs over sentences, which could instead lead to a computational cost that grows with $N_1N_2D$. Additional sentence-level query and key mappings in Eq. \ref{eq:query_mapping} and Eq. \ref{eq:key_mapping} also incur a large computational cost.

\subsubsection*{Model-free Approximator} 
Lastly, we re-visit the sentence-level attention in Eq. \ref{eq:sent_level_attn}, which have been approximated by Eq. \ref{eq:sent_level_approximation} and the model-based approximator. It is a challenge to attempt via a model-free algebraic approximation, which does not require any training and has little additional inference-time cost. We examined various forms, and we present one model-free approach as well as experimental results in Appendix \ref{sec:model_free_appendix}, but the current form has worse summarization performance than our model-based approach.

\section{Related Work}
The discrepancy between low/moderate attention weight sparsity and good sparse approximation could be because a considerable amount of the attention weight is assigned to special tokens, e.g. `{\tt .}' in all sentences, but their vector norm is small, which was observed in \citet{kobayashi-etal-2020-attention}.

The sparse attention (Eq. \ref{eq:subset_attention_hat}) with ideal selection (Eq. \ref{eq:idealI}) can be considered as \textit{content selection}, which has been shown to improve summarization \cite{gehrmann-etal-2018-bottom, hsu-etal-2018-unified}. Recently, head-wise masks are applied to encoder-decoder attention at inference time, and a performance improvement is reported \cite{cao2021attention}.

\citet{voita-etal-2019-analyzing} observed that heads are redundant, and \citet{clark-etal-2019-bert} found that a head in BERT rarely attends to several consecutive tokens. Based on these, \citet{huang2021effattn} applies a stride pattern in the encoder-decoder attention, reducing its cost by a factor of the stride size, and this method is likely complementary to our work. 

\section{Conclusion}
We show that the computational cost of the transformer decoder becomes more significant at inference time. Towards reducing this cost, first, we show that there is sparsity in the encoder-decoder attention that allows us to reduce the computational cost with a minimal degradation. Second, we partition the sentence-level attention score, and we augment the standard decoder by adding a neural network to approximate the attention over sentences, allowing sentence selection. We show that the summarization performance of our approach converges to that of the full attention baseline, while switching the complexity from $\mathcal{O}(MN_1N_2)$ to $\mathcal{O}(MN_1+k_wMrN_2+k_eN_1N_2)$.

\section*{Acknowledgments}
This paper reports on research supported by ALTA institute, Cambridge Assessment English, University of Cambridge, and Cambridge International \& St John’s College Scholarship. Thanks to Yiting Lu for interesting discussions. Thanks to the anonymous reviewers for their helpful comments.

\newpage
\bibliography{anthology,custom}
\bibliographystyle{acl_natbib}

\newpage
\appendix
\section{Reproducibility Details}
\label{sec:reproducibility}
\subsection{Models}
\textbf{BART/LoBART}: We use the HuggingFace's implementation \cite{wolf-etal-2020-transformers}, including BART models fine-tuned to CNNDM\footnote{\url{https://huggingface.co/facebook/bart-large-cnn}} and XSum\footnote{\url{https://huggingface.co/facebook/bart-large-xsum}}. We take LoBART from \citet{manakul2021_longspan}, including LoBART(4k)+MCS fine-tuned to Podcast and arXiv. MCS is the multitask content selection system for handling the Podcast/arXiv input documents that exceed 4096 words.

\vspace{4pt}
\noindent \textbf{Modified Architecture}: As shown in Fig. \ref{fig:modified_architecture}, the modified architecture consists of a sentence encoder and a sentence-level encoder-decoder attention. The sentence encoder, which approximates $f_2(.)$ in Eq. \ref{eq:sent_level_approximation}, is a two-layer bi-directional GRU \cite{cho-etal-2014-learning} with hidden dimension of 1024. The sentence-level encoder-decoder attention has linear mapping weights $\mathbf{W}^{\tt Q}$ (query) and $\mathbf{W}^{\tt K}$ (key), and both weights have the same dimension as and are initialized from their corresponding word-level encoder-decoder attention's linear mapping weights in the base model. The total number of additional parameters is 58.8M.

\subsection{Data}
\textbf{CNNDM}: We follow the standard train/valid/test split of 287,113/13,368/11,490. 

\vspace{4pt}
\noindent \textbf{XSum}: We follow the standard train/valid/test split of 204,045/11,332/11,334. 

\vspace{4pt}
\noindent \textbf{Spotify Podcast}: We follow the data processing and split in \citet{jones_trec2020}, resulting in train/valid/test splits of 60,415/2,189/1,027. 

\vspace{4pt}
\noindent\textbf{arXiv}: We follow the standard train/valid/test split of 203,037/6,436/6,440. 

\vspace{4pt}
Our data processing is based on the byte-pair-encoding (BPE) tokenizer same as the BART-large tokenizer, and we use the NLTK toolkit for sentence splitting. **In Fig. \ref{fig:summary_performance} and Fig. \ref{fig:varying_r}, to reduce computational cost, we use first 2,000 samples of each test set, except Podcast which contains less than 2,000 samples.

\begin{table}[!h]
  \centering
  \begin{tabular}{rrrrrr}
    \toprule
    Dataset &$N$ &$N_1$  &$M$ &$N/M$   \\
    \midrule
    CNNDM     &870 &28.7   &67.4  &14.2 \\
    XSum      &489 &17.4   &27.9  &18.2 \\
    Podcast  &5727 &330.8  &86.6 &143.9 \\ 
    arXiv    &8584 &237.0 &364.9  &45.3 \\
    \bottomrule
  \end{tabular}
  \caption{Data statistics (average over corpus). $N$=input length, $N_1$=\#sentences, $M$=summary length.}
  \label{tab:data_statistics}
\end{table}

\subsection{Training and Inference}
\label{sec:training_appendix}
We use PyTorch \cite{pytorch} in our experiments. All training experiments use the Adam optimizer \cite{kingma2014adam} with $\beta_1$=0.9, $\beta_2$=0.999, and the learning rate is:
\begin{equation*}
    \text{lr} = 0.002 \times \text{min}(\text{step}^{-0.5}, \text{step}\times\text{warmup}^{-1.5})
\end{equation*}
where we use 20,000 warmup steps. In all experiments, we set batch size to 1, and gradient accumulation to 2 steps. We evaluate the training loss on the validation set every 20,000 steps, and stop the training if the validation loss does not improve 3 times. All training experiments converged within 1 epoch. All experiments were carried out in 32-bit precision on either one V100 (32GB) GPU, or one RTX 2080Ti (11GB) GPU. 

At inference time, we use the standard setting: beam search of width 4, and length penalty of 2.0 \cite{wu2016google} for all experiments. The ROUGE \cite{lin-2004-rouge} scoring tool is \texttt{pyrouge}.\footnote{\url{https://pypi.org/project/pyrouge/}}

\subsection{Multi-head attention}
\label{section:multihead_appendix}
In all of the equations and expressions in the paper, we omit the heads for simplicity. Both BART and LoBART models have 16 heads. In Fig. \ref{fig:attention_retained}, we average $\alpha_{m,i}^{\tt s}$ over heads, before the summation.  When computing an uncertainty measure such as entropy $\text{H}(.)$ or KL-divergence $\text{KL}(.)$, we compute the measure for each head separately and take the average. In obtaining $\mathcal{I}^r_m$, we average $\alpha_{m,i}^{\tt s}$ over heads, before the top-$r$ operation, i.e. all heads get assigned the same subset of sentences, but the differences are across layers and decoding timesteps.

\subsection{KL Loss and Integrated Training}
The target ${\alpha}_{m,i}^{{\tt s}}$ is re-normalized to encourage higher sparsity as follows:
\begin{equation}
    \boldsymbol{\alpha}_{m}^{{\tt s}} \leftarrow \text{softmax}\left(\frac{\log(\boldsymbol{\alpha}_{m}^{{\tt s}})}{T}\right)
\end{equation}
where temperature $T$ is set to 0.5. For integrated training, we set $\lambda$ in $\mathcal{L}_{\tt I}$ to 0.2. We initialize integrated training experiments from KL-only models. 

\section{Time Analysis}
\label{sec:time_appendix}
For each mode in Table \ref{tab:decoder_cost}, we take 6 samples of $M$ and the average time of 100 iterations. Curve fitting yields R-squared of at least 0.994. 

Computational time as function of $M$ and $N$ is ${\tt time}=c_1+c_2M+c_3N+c_4MN+c_5M^2+c_6N^2$. The coefficients are obtained by a least-squares regression, e.g. $\mathbf{c}^*= (\mathbf{P}^T\mathbf{P})^{-1}\mathbf{P}^T \mathbf{t}$ where $\mathbf{P}$ is the matrix of $M,N$ associated with the coefficients, and $\mathbf{t}$ contains the time measures. We collect 30 samples of the average time of 100 F+B passes, spanning $N\in[256,1024]$ and $M\in[50,300]$. The normalized coefficients are: $c_1=1.00, c_2=3.78\times10^{-3}, c_3=3.15\times10^{-3}, c_4=1.47\times10^{-6}, 
c_5=7.26\times10^{-7}, c_6=7.79\times10^{-7}$. Because $c_4 \approx 2c_5$ and $N>M$, the enc-dec attention cost is greater than the decoder self attention cost.

\section{Sensitivity to Random Selection}
\label{section:random_selection_appendix}
\vspace{-0.2cm}
\begin{table}[!h]
  \centering
  \scalebox{0.9}{
  \begin{tabular}{rccc}
    \toprule
    System    &R1 &R2 &RL  \\
    \midrule
   Vanilla ($\gamma=0.0$) &39.06 &14.32 &36.07  \\
   $\mathcal{L}_{\tt A}$-tuned ($\gamma=0.1$) &28.43 &7.72  &24.00  \\
   $\mathcal{L}_{\tt A}$-tuned ($\gamma=1.0$) &21.38 &4.22  &17.69  \\
    \bottomrule
  \end{tabular}}
  \caption{Impact of sparsity on the sensitivity to random selection based on CNNDM with $r=5$.}
  \label{tab:random_selection}
\end{table}
\vspace{-0.45cm}
\section{Model-free Approximation for Eq. \ref{eq:sent_level_attn}}
\label{sec:model_free_appendix}
Since $\alpha^{\tt s}_{m,i} = \frac{1}{Z_m} \sum_{j=1}^{J_i} \exp (\mathbf{q}_m \cdot \mathbf{k}_{i,j})$ is used to rank input sentences, the normalization term, $Z_m$, can be dropped. The encoder-decoder attention has $\mathcal{O}(MN_1N_2)$ complexity because $\mathbf{q}_m \cdot \mathbf{k}_{i,j}$ is computed for every $m$ and $(i,j)$ pair. Hence, if we can group $\mathbf{k}_{i,j}$ into sentences, the complexity could potentially be reduced. We try the following approximation of unnormalized $\alpha^{\tt s}_{m,i}$:
\vspace{-8pt}
\begin{align}
    \sum_{j=1}^{J_i} \exp &(\mathbf{q}_m \cdot \mathbf{k}_{i,j}) \nonumber \\
    \vspace{-8pt}
    &= \sum_{j=1}^{J_i} \prod_{d=1}^D\exp ({q}_{m,d}{k}_{i,j,d}) \label{eq:phi_exact_prod} \\
    \vspace{-1pt}
    &\approx \sum_{j=1}^{J_i} \sum_{d=1}^D\exp ({q}_{m,d}{k}_{i,j,d}) \label{eq:phi_first_approx} \\
    \vspace{-1pt}
    &\approx \sum_{j=1}^{J_i} \sum_{d=1}^D\phi ({q}_{m,d})\phi({k}_{i,j,d}) \label{eq:phi_second_approx} \\
    \vspace{-2pt}
    &= \phi (\mathbf{q}_{m}) \cdot \left(\sum_{j=1}^{J_i} \phi(\mathbf{k}_{i,j}) \right) \label{eq:phi_final}
\end{align}
where $d=\{1,...,D\}$ is the hidden dimension, and $\phi(.) = \text{ELU}(.)+1$ (or other form such as $\exp$ and ReLU). The model-free method reduces complexity from $\mathcal{O}(MN_1N_2)$ to $\mathcal{O}(MN_1 + k_wMrN_2 + k_eN_1N_2)$ where $k_e$ is now much smaller compared to the model-based approach. Based on Eq. \ref{eq:phi_final}, we provide model-free results in Table \ref{tab:phi_experiment}.

\begin{table}[!h]
  \centering
  \scalebox{0.9}{
  \begin{tabular}{rccc}
    \toprule
    Selection Method   &R1 &R2 &RL  \\
    \midrule
     Ideal (Eq.\ref{eq:idealI})    &43.94 &20.82 &40.81 \\
     Best Model-based &43.72 &20.40 &40.70 \\
     Random    &39.06 &14.32 &36.07  \\
     \midrule
    Model-free (Eq.\ref{eq:phi_final}) &40.01 &17.28 &36.95 \\
    \bottomrule
  \end{tabular}}
  \caption{Model-free results on CNNDM ($r=5$).}
  \label{tab:phi_experiment}
\end{table}
Our model-free approach is better than the random selection baseline, but it is significantly worse than both the ideal selection baseline and the model-based approach. The reasons for this poor performance are: (i) the approximation from (\ref{eq:phi_exact_prod}) to (\ref{eq:phi_first_approx}) requires the following condition to be true: $A_1A_2 > B_1B_2 \rightarrow A_1+A_2 > B_1+B_2$, so it is inaccurate when the values are not in a similar range; (ii) the approximation from (\ref{eq:phi_first_approx}) to (\ref{eq:phi_second_approx}) is inaccurate for non-positive values. In conclusion, this experiment investigates an alternative challenging method, which would not require any training and would be computationally cheaper at inference time. Although the current algebra does not work well, we hope that our initial study might draw more interests into this type of model-free approach to exploit the sentence structure in seq2seq tasks such as abstractive summarization.

\section{Word-level and Sentence-level Attention Weight Plots}
\label{sec:attention_plot_appendix}
\begin{figure*}[!t]
    \begin{subfigure}[b]{0.49\linewidth}
    \centering
      \includegraphics[width=\linewidth,keepaspectratio]{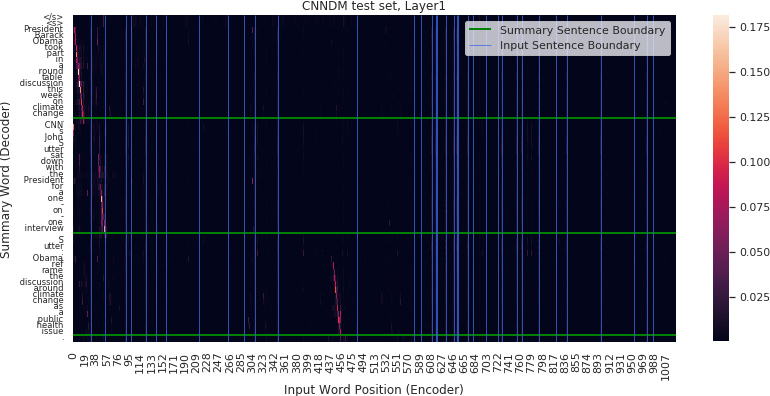}
    \caption{Word-Level}
    \label{fig:attention_plot_word_cnndm}  
    \end{subfigure}
        \hfill
    \begin{subfigure}[b]{0.49\linewidth}
    \centering
      \includegraphics[width=\linewidth,keepaspectratio]{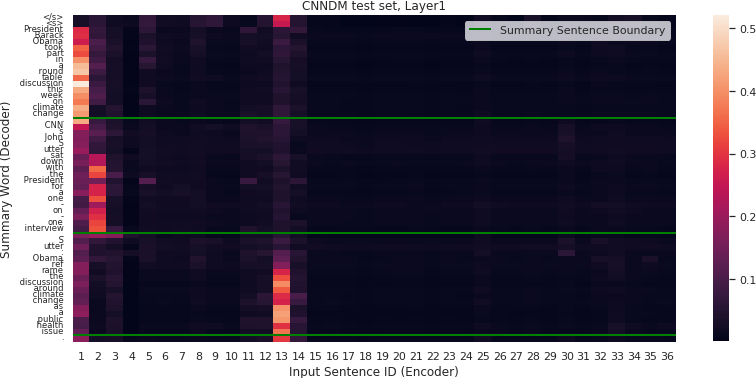}
    \caption{Sentence-Level ($\alpha_{m,i}^{\tt s}$)}
    \label{fig:attention_plot_sent_cnndm} 
    \end{subfigure}
    \caption{Example of BART's encoder-decoder attention evaluated on CNNDM test set.}
    \label{fig:attention_plot_cnndm}    
\end{figure*}

\begin{figure*}[!t]
    \begin{subfigure}[b]{0.49\linewidth}
    \centering
      \includegraphics[width=\linewidth,keepaspectratio]{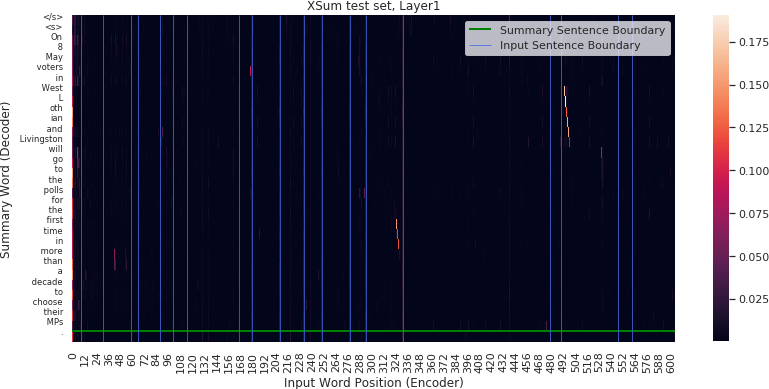}
    \caption{Word-Level}
    \label{fig:attention_plot_word_xsum}  
    \end{subfigure}
        \hfill
    \begin{subfigure}[b]{0.49\linewidth}
    \centering
      \includegraphics[width=\linewidth,keepaspectratio]{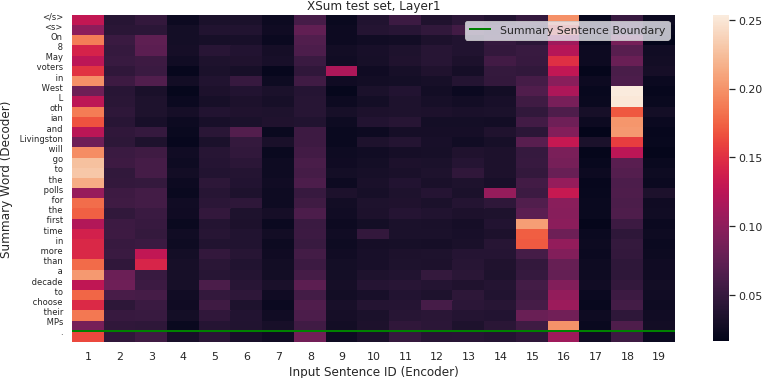}
    \caption{Sentence-Level ($\alpha_{m,i}^{\tt s}$)}
    \label{fig:attention_plot_sent_xsum} 
    \end{subfigure}
    \caption{Example of BART's encoder-decoder attention evaluated on XSum test set.}
    \label{fig:attention_plot_xsum}    
\end{figure*}

\begin{figure*}[!t]
    \begin{subfigure}[b]{0.49\linewidth}
    \centering
      \includegraphics[width=\linewidth,keepaspectratio]{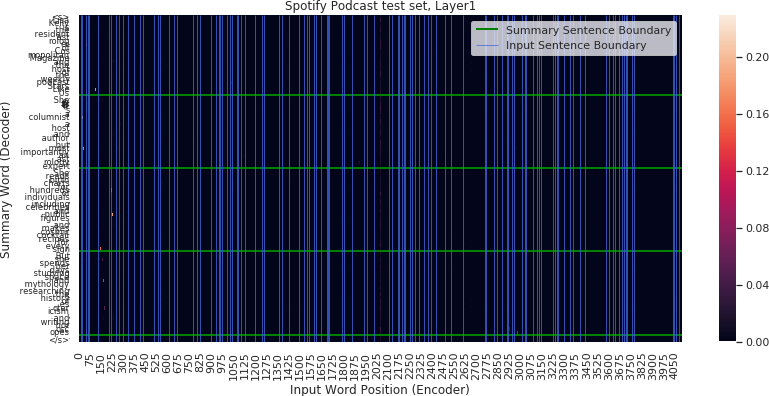}
    \caption{Word-Level}
    \label{fig:attention_plot_word_podcast}  
    \end{subfigure}
        \hfill
    \begin{subfigure}[b]{0.49\linewidth}
    \centering
      \includegraphics[width=\linewidth,keepaspectratio]{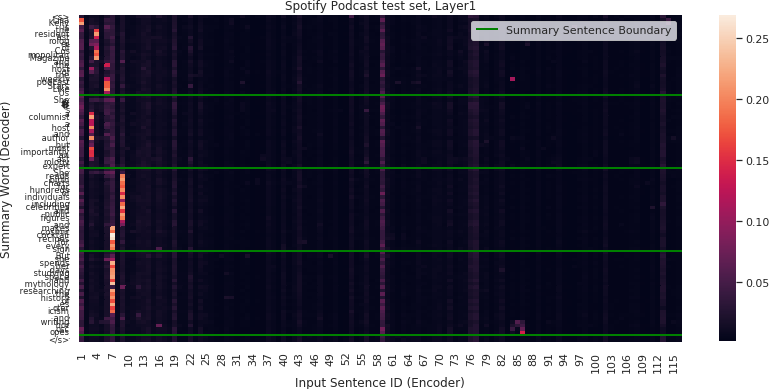}
    \caption{Sentence-Level ($\alpha_{m,i}^{\tt s}$)}
    \label{fig:attention_plot_sent_podcast} 
    \end{subfigure}
    \caption{Example of LoBART's encoder-decoder attention evaluated on Podcast test set.}
    \label{fig:attention_plot_podcast}    
\end{figure*}

\begin{figure*}[!t]
    \begin{subfigure}[b]{0.49\linewidth}
    \centering
      \includegraphics[width=\linewidth,keepaspectratio]{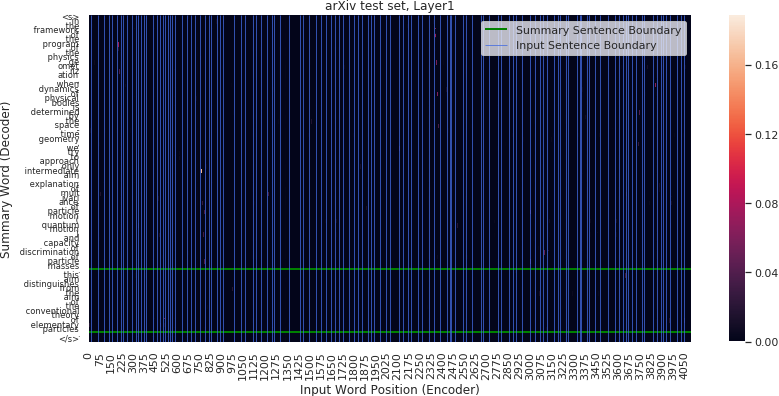}
    \caption{Word-Level}
    \label{fig:attention_plot_word_arxiv}  
    \end{subfigure}
        \hfill
    \begin{subfigure}[b]{0.49\linewidth}
    \centering
      \includegraphics[width=\linewidth,keepaspectratio]{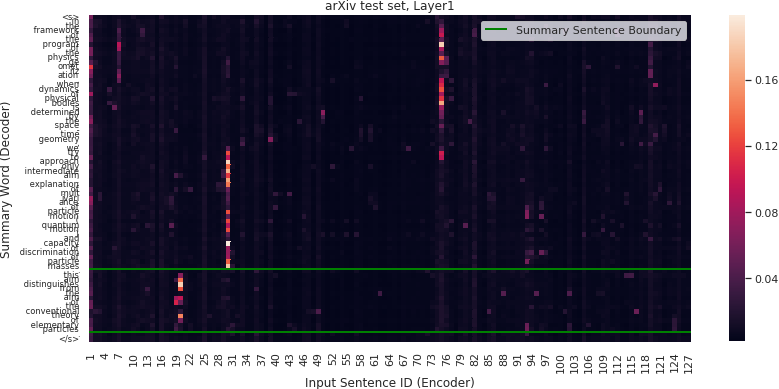}
    \caption{Sentence-Level ($\alpha_{m,i}^{\tt s}$)}
    \label{fig:attention_plot_sent_arxiv} 
    \end{subfigure}
    \caption{Example of LoBART's encoder-decoder attention evaluated on arXiv test set.}
    \label{fig:attention_plot_arxiv}    
\end{figure*}

Constraining the encoder-decoder attention to $r$ sentences is motivated by the observations of sentence-level attention (in Fig. \ref{fig:attention_plot_sent_cnndm}, \ref{fig:attention_plot_sent_xsum}, \ref{fig:attention_plot_sent_podcast}, \ref{fig:attention_plot_sent_arxiv}). Note that we average over all heads for the plots. 

For instance, Fig. \ref{fig:attention_plot_cnndm} shows that the decoder attends particularly to input sentences \#1,\#2,\#13 in the summary generation. Compared to Fig. \ref{fig:attention_plot_cnndm}, Fig. \ref{fig:attention_plot_xsum} shows a wider spread of the attention over sentences in a more abstractive task. When using LoBART, Fig. \ref{fig:attention_plot_podcast} and Fig. \ref{fig:attention_plot_arxiv} show a similar trend of the sparsity to BART scenarios. These figures also explain Fig. \ref{fig:attention_retained_vanilla} (Section \ref{sec:encdec_attn}) that $\sum_{i}^{\mathcal{I}^r_m} \alpha_{m,i}^{\tt s}$ is only low/moderate because most sentences get assigned some attention weights, despite being non-salient. 


\end{document}